\documentclass{article}

\usepackage{arxiv}

\usepackage[utf8]{inputenc} % allow utf-8 input
\usepackage[T1]{fontenc}    % use 8-bit T1 fonts
\usepackage{hyperref}       % hyperlinks
\usepackage{url}            % simple URL typesetting
\usepackage{booktabs}       % professional-quality tables
\usepackage{amsfonts}       % blackboard math symbols
\usepackage{nicefrac}       % compact symbols for 1/2, etc.
\usepackage{microtype}      % microtypography
\usepackage{lipsum}         % Can be removed after putting your text content
\usepackage{graphicx}
\usepackage{natbib}
\usepackage{doi}
\usepackage{comment}

\usepackage{graphicx}%
\usepackage{multirow}%
\usepackage{amsmath,amssymb,amsfonts}%
\usepackage{amsthm}%
\usepackage{mathrsfs}%
\usepackage[title]{appendix}%
\usepackage{xcolor}%
\usepackage{textcomp}%
\usepackage{manyfoot}%
\usepackage{booktabs}%
\usepackage{algorithm}%
\usepackage{algorithmicx}%
\usepackage{algpseudocode}%
\usepackage{listings}%
\usepackage{longtable}%

\title{An inclusive review on deep learning techniques and their scope in handwriting recognition}

\newif\ifuniqueAffiliation
\uniqueAffiliationtrue

\ifuniqueAffiliation % Standard variant of author block
\author{ \href{https://orcid.org/0000-0000-0000-0000}{\includegraphics[scale=0.06]{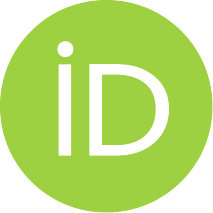}\hspace{1mm}Sukhdeep Singh}\\
	D.M. College (Affiliated to Panjab University, Chandigarh)\\
	Moga, Punjab, India\\
	\texttt{sukha13@ymail.com} \\
	\And
	\href{https://orcid.org/0000-0000-0000-0000}{\includegraphics[scale=0.06]{orcid.pdf}\hspace{1mm}Sudhir Rohilla} \\
	Department of Computer Science\\
	Gopichand Arya Mahila College\\
	Abohar, Punjab, India\\
	\texttt{rohilla2209@gmail.com} \\
	\And
	\href{https://orcid.org/0000-0000-0000-0000}{\includegraphics[scale=0.06]{orcid.pdf}\hspace{1mm}Anuj Sharma} \\
	Department of Computer Science and Applications\\
	Panjab University, Chandigarh, India\\
	\texttt{anujs@pu.ac.in} \\
}
\else
\usepackage{authblk}

\newbox{\orcid}\sbox{\orcid}{\includegraphics[scale=0.06]{orcid.pdf}} 
\author[1]{%
	\href{https://orcid.org/0000-0000-0000-0000}{\usebox{\orcid}\hspace{1mm}Sukhdeep Singh\thanks{\texttt{sukha13@ymail.com}}}%
}
\author[2]{%
	\href{https://orcid.org/0000-0000-0000-0000}{\usebox{\orcid}\hspace{1mm}Sudhir Rohilla\thanks{\texttt{rohilla2209@gmail.com}}}%
}
\author[3]{%
	\href{https://orcid.org/0000-0000-0000-0000}{\usebox{\orcid}\hspace{1mm}Anuj Sharma\thanks{\texttt{anujs@pu.ac.in}}}%
}
\affil[1]{D.M. College (Affiliated to Panjab University, Chandigarh),
	Moga, Punjab, India}
\affil[2]{Department of Computer Science, Gopichand Arya Mahila College, Abohar, Punjab, India}
\affil[3]{Department of Computer Science and Applications, Panjab University, Chandigarh, India}
\fi

\renewcommand{\shorttitle}{\textit{arXiv} Template}

\hypersetup{
pdftitle={A template for the arxiv style},
pdfsubject={q-bio.NC, q-bio.QM},
pdfauthor={David S.~Hippocampus, Elias D.~Striatum},
pdfkeywords={First keyword, Second keyword, More},
}

\begin{document}
\maketitle

\begin{abstract}
Deep learning expresses a category of machine learning algorithms that have the capability to combine raw inputs into intermediate features layers. These deep learning algorithms have demonstrated great results in different fields. Deep learning has particularly witnessed for a great achievement of human level performance across a number of domains in computer vision and pattern recognition. For the achievement of state-of-the-art performances in diverse domains, the deep learning used different architectures and these architectures used activation functions to perform various computations between hidden and output layers of any architecture. This paper presents a survey on the existing studies of deep learning in handwriting recognition field. Even though the recent progress indicates that the deep learning methods has provided valuable means for speeding up or proving accurate results in handwriting recognition, but following from the extensive literature survey, the present study finds that the deep learning has yet to revolutionize more and has to resolve many of the most pressing challenges in this field, but promising advances have been made on the prior state of the art. Additionally, an inadequate availability of labelled data to train presents problems in this domain. Nevertheless, the present handwriting recognition survey foresees deep learning enabling changes at both bench and bedside with the potential to transform several domains as image processing, speech recognition, computer vision, machine translation, robotics and control, medical imaging, medical information processing, bio-informatics, natural language processing, cyber security, and many others.
\end{abstract}

% keywords can be removed

\section{Introduction}\label{sec1}
The intelligent act with synthesis and analysis of computational agents represents Artificial Intelligence (AI). Here, an agent is who completes the signed goal with various learning techniques and training of data. The agent when computationally represented, it is called computational agent \cite{poole10, rich10}. The artificial intelligence has made our life very exciting with state-of-the-art research in this area. However, the research in AI regularly demands new paradigms that could further help in error-free AI systems. The AI has many areas of research such as machine learning, data mining, intelligent tutoring, case-based reasoning, multi-agent planning, scheduling, uncertain reasoning, natural language understanding and translation, vision, virtual reality, games, robotics and other topics \cite{richter19, chen20, nilsson10, goodrich07, buczak16, arash10, bengio13, david18, gustavo18, zoubin15, cchi16}. The one of today's popular research fields in AI is Machine Learning (ML). The machine learning mainly includes intelligent system development using training of data. Therefore, the ML based system model developed with train data further decides the nature of future data as test data. The common techniques of machine learning are data understanding, regression, clustering, classification, dimension reduction, deep learning, big data, online learning etc \cite{donald89, meci19, bishop06, chapelle06, collobert11, du11, freund97, grira04, guyon06, lecun15, pedregosa11, vapnik98}. Here, each ML technique offers uniqueness in terms of data handling, feature computation and respective output. Data understanding is data normalization and processing of data; regression is promising statistical area to understand continuous type of data; clustering allow class formation of data; classification distinguish data for various classes; dimension reduction reduce feature size of data and retain useful information for data; deep learning is promising data classification and understanding area; big data is dedicated to handle large amount of data using established scientific methods; online learning refers to handle data as it comes and not in conventional batch mode. The recent research in ML suggests that deep learning is one of the promising techniques to achieve high accuracy results. Therefore, deep learning studied by many scientists in recent past and it has been observed that suitable literature of deep learning always helps for readers working in this area. Especially, suitable review of deep learning two popular methods as Convolutional Neural Networks (CNN) and Recurrent Neural Networks (RNN) will be meaningful from research and development point of view \cite{hamid14, cho14, zhao17}. 
\par In machine learning, representational learning allow automatically appropriate way to represent data. Deep learning is one such technique which includes representation learning feature. The progression of learned transformation helps deep learning to achieve automatic representation of data. This has been realised in recent past that deep learning-based applications achieved promising results for large amount of data problem, which were not possible to automate data representation before deep learning techniques. The popular deep learning model and its practical implementation based quintessential example is deep network. These deep networks are inspired from common machine learning technique as multilayer perceptron which is a common neural network algorithm. The depth of deep network as a computer program is organised in such a way that each layer meant for a specific purpose and it could recall other layers' computer memory when working in parallel. This way, the network with more depth allows many instructions in a sequence. The two approaches of deep networks as CNN and RNN proved state-of-the-art results in recent past. Further, LSTM and BLSTM are two particular forms of RNN. The two competitive techniques as CNN and RNN share many commonalities in working, however, they do differ in suitability of various types of data. Interestingly, hybrid technique using CNN and RNN is another feasible technique with promising results.

\par In literature, we find many studies that include the working of CNN and RNN. However, overall working of CNN and RNN with their architecture, mathematical formulation and implementation from review point of view needs attention and is presented in this paper. This paper has been presented with updated literature information and focused on theory as well as the implementation of deep networks. This review is aiming to answer the question of the working of deep networks, CNN and RNN, and their performance in handwriting recognition in recent past with the availability of latest system configurations. As a result, we highlight in summary the following contributions as: (i) deep networks outperform in handwriting recognition, (ii) the CNN and RNN results in state-of-the-art results for real life challenging datasets, (iii) the architecture of deep networks offers enormous scope to researchers to enhance its architecture, and further offer many areas of research to improve deep networks. Moreover, this paper also presents general observations of deep networks and their applications in handwriting recognition based on suitable findings from literature work. In this manner, it has been analysed that deep networks solely justify deep learning representative to real life handwriting recognition applications. This analysis is based on the results discussed in this paper using benchmark datasets with variants of deep learning approaches.

\par This paper maintains continuity as deep networks, CNN and RNN, architectures and DL results for handwriting recognition in a sequence for better understanding from the reader's point of view. The rest of the paper is organized as follows. This article describes the deep networks' fundamentals and evolution in section 2. The section 3 demonstrates different deep learning architectures. The section 4 presents existing literature results using deep networks as CNN and RNN in handwriting recognition. The section 6 presents the general observations based on literature finding for CNN and RNN. The last section 7 concludes this article with findings and scope of future work.

\section{Deep Forward Networks}\label{sec2}

\par Deep learning quintessential are forward networks and forward networks are commonly called deep forward networks. The forward networks are also known as multi-layer perceptron as it follows multiple layers architecture using perceptron concept. The CNN and RNN are the emergent variants of deep learning framework based on neural networks. Therefore, prior understanding of neural networks is important in this review to know how neural networks work and CNN or RNN based on neural networks. Before the introduction of deep neural networks, this section illustrates the evolution of neural networks with common benchmark algorithms of literature including most preferred backpropagation algorithm.
\par In early works, McCulloch and Pitts neuron was designed in 1943, it mainly included the combination of logic functions with the concept of threshold \cite{culloch43}. The Hebb network was designed in 1949, which included two active neurons simultaneously with their strong inter-connections \cite{hebb49}. One major contribution was perceptron model in 1958 by Rossenblatt, which included weights in connection path with their adjustment \cite{blatt58}. In 1960, Adaline network was built with the ability to reduce difference between net input weights and out weights and resulted in minimizing the mean error rates \cite{windrow60}. One major development noticed for unsupervised learning in 1982, Kohonen introduced Kohonen self-organizing maps where inputs were clustered together to form output neurons \cite{kohonen82}. This work was among initial findings to understand neural networks in supervised and unsupervised areas. One such study with fixed weights was done for Hopfield networks to act as associative memory nets \cite{hopfield82}. In 1986, a complete neural network algorithm, with forward and backward ability to update weights and based on Multi-Layer Perceptron (MLP), was introduced as backpropagation algorithm \cite{hart86}. This backpropagation algorithm was a complete multi-layer perceptron technique. This architecture included input, hidden and output layers, forward moves to update weights and backward moves to improve weights with propagated error information at output unit in each iteration. After backpropagation, neural networks witnessed many improvements subject to the nature of problem. Few networks were adaptive resonance theory, radial basis functions, neocognition until 1990 \cite{art88, bro88, fuku90}. The main development with state-of-the-art results were reported with MLP until the introduction of convolution networks in 2000.
\par
\begin{figure}
	\centering
	\includegraphics[width=0.60\textwidth]{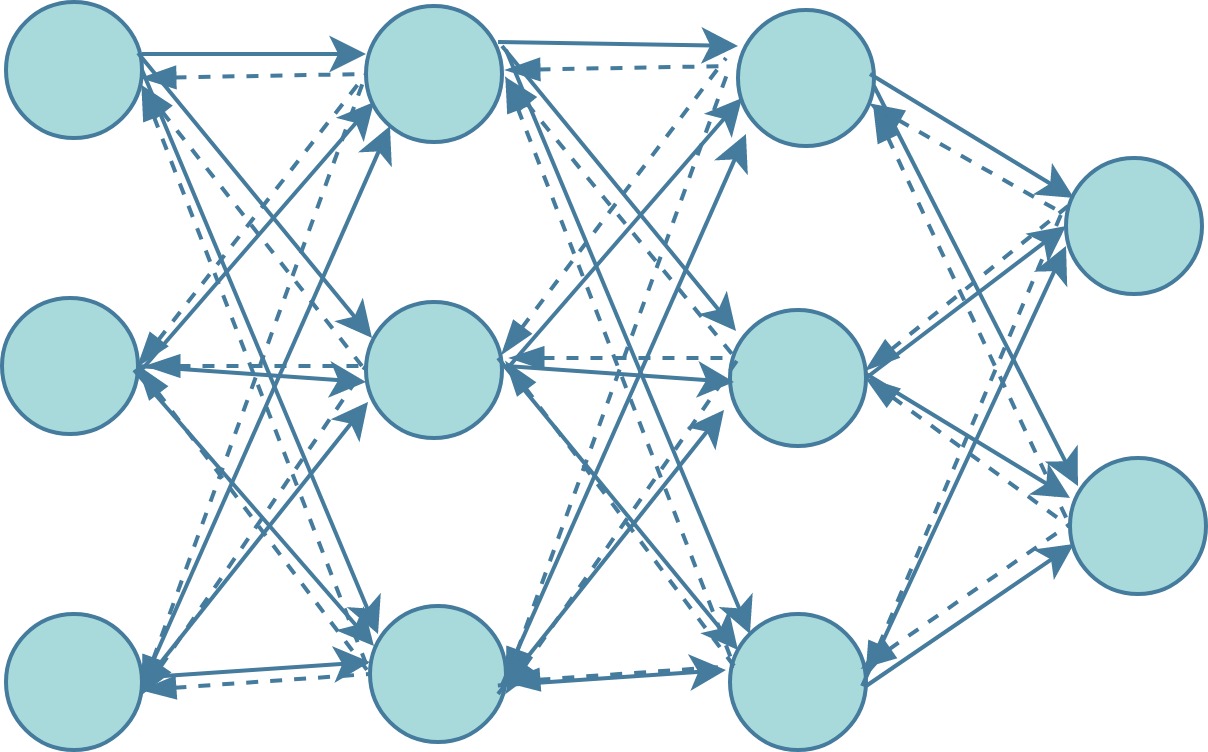}
	\caption{Overview of multi layer perceptron technique}
	\label{MLPfig1}       % Give a unique label
\end{figure}

 The MLP or deep forward network is a massively parallel distributed processor made up of simple processing units, which has a natural propensity for storing experiential knowledge and making it available for use \cite{haykin98}. It is a directed graph consisting of nodes with interconnecting synaptic and activation links with main properties as: each neuron is represented by a set of linear synaptic links with bias, and a possibly nonlinear activation link; the synaptic links of a neuron weight to their respective input signals; the weighted sum of the input signals defines the induced local field of the current neuron; thus, the activation link squashes the induced local field of the neuron to produce an output.  The presence of one or more layers between input and output layers are called hidden layers and nodes of corresponding layers are hidden neurons. This enhances system learning capability and is referred as multi-layer networks and results in MLP. The major characteristics of MLP include: the model of each neuron in network includes nonlinear activation function; the network includes multiple hidden layers that enable network to learn complex tasks by progressively extracting more meaningful features by minimizing errors at output layer; the network exhibits high degree of connectivity determined by synapses of network and a change in network require change in synaptic connections or their weights. The working of MLP has been presented in \ref{fig1}, where input layer includes three nodes, two hidden layers with three nodes each and output layer with two nodes. In figure \ref{MLPfig1}, forward move is shown with connected lines and backward moves with dotted lines. The input layer includes initialized values that are processed with weight vectors and each hidden layer node is updated in forward manner. The output layer computes final value and error is computed subject to target value against output value. Therefore, error values decide to move backward in order to minimize error values. This results in many iterations until the expected value of error is achieved. This way, it includes the computation of the function signal appearing at the output neuron, a continuous nonlinear function of input signal and synaptic weight associated with that neuron. Also, the computation of an estimate of gradient vector which is needed for backward pass of the network.

\par The MLP algorithm includes three major steps as forward computation, backward computation and weight updating. The working of these steps has been presented with an instance of one hidden layer to another consecutive hidden layer for neuron $j$ to neuron $k$. The details of MLP including derivations have been studied extensively in literature. According to \cite{haykin98}, for an MLP, the first step as forward computation results in output as $y_{j}^{(l)}$ for neuron $j$ in layer $l$ using:

\begin{center}
	$v_{j}^{(l)}(n)=\sum_{i=0}^{m_{0}}w_{ji}^{(l)}(n)y_{i}^{(l-1)}(n)$;\\
	$y_{j}^{(l)}=\phi_{j}(v_{j}(n))$
\end{center}

The forward computation is performed in forward direction until the output layer neuron value is not computed. In backward computation, gradient value is computed for neuron $j$ in hidden layer $l$ as:
\begin{center}
	$\delta_{j}^{(l)}(n)=e_{j}^{(L)}(n)\phi_{j}^{'}(v_{j}^{L}(n))$, for neuron $j$ in output layer $L$;\\
	$\delta_{j}^{(l)}(n)=\phi_{j}^{'}(v_{j}^{l}(n))\sum_{k}\delta_{k}^{(l+1)}(n)w_{kj}^{(l+1)}(n)$\\
\end{center}
For one forward and backward computation, it completes one iteration and updates the weight for next iteration. The adjustment of the synaptic weights of network in layer $l$ is computed as:        
\begin{center}
	$w_{ji}^{(l)}(n+1)=w_{ji}^{(l)}(n)+\alpha[w_{ji}^{(l)}(n-1)]+\eta\delta_{j}^{(l)}(n)y_{i}^{(l-1)}(n)$
\end{center}
In above three steps of MLP, $w_{ji}$ is the weight from $i$ to $j$; $v_{j}^{(l)}(n)$ is neuron $j$ in layer $l$; $y_{j}^{(l)}$ is output of neuron $j$ in layer $l$; $\delta_{j}^{(l)}(n)$ is gradient for neuron $j$ in layer $l$; $e_{j}^{(L)}(n)$ is error signal at layer $L$ for neuron $j$; $\phi_{j}^{'}$ is differentiation of activation function; $\eta$ and $\alpha$ are the learning rate and momentum constant respectively.

\section{Deep Learning Architecture}\label{sec3}

\par The pattern recognition field has started using DL architectures extensively and image recognition \cite{krizhevsky2012imagenet} \cite{Szegedy_2015_CVPR} has also attained good performance for recognizing faces \cite{taigman2014deepface}, text recognition \cite{simard2003best} \cite{Ciresan2011} \cite{ciregan2012multi} \cite{wang2012end} \cite{goodfellow2013multi} and estimation of human poses \cite{tompson2014joint}. Deep learning, either uses deep architectures of learning or hierarchical learning approaches, is a class of machine learning developed mostly after 2006. The DL architectures have basically made alteration of the traditional form of pattern recognition and have contributed a major development in various handwriting recognition tasks too. The traditional ML approaches perform better for lesser amounts of input data. When the data size increases beyond a certain limit, the performance of traditional machine learning approaches becomes steady, whereas deep learning performance increases with respect to the increment of data size. The key breakthrough of DL is that these models can perform feature extraction and classification automatically. Deep learning architectures with more than one hidden layer are referred to as multilayer perceptron. These architectures include deep feed forward neural network, CNN, RNN, LSTM and the deep generative models as deep belief networks, deep Boltzmann machines, generative adversarial networks \cite{lecun2015deep} \cite{Goodfellow-et-al-2016}. Deep learning architectures include to learn patterns in data, mapping of input function to outputs and many more, and it can be attained with specialized architectures only. Among DL architectures, the CNN and RNN: LSTM and BLSTM are the most commonly used vital architectures. LSTM and BLSTM are particularly designed for data in sequential form, have been used in the studies of pattern and handwriting recognition. One of the key problems with the RNN deep network is that the hidden layers are influenced by the input layer and consequently the output layer goes on decaying as it cycles through the network's recurrent connections, and this problem is known as vanishing gradient problem \cite{Bengio1994}. Such problem becomes a cause for an incomplete range of contextual information access by RNN, and the contextual information cannot be retained for a longer period of time by an architecture of RNN.

\subsection{Convolutional Neural Network}
\par Among different models of DL, the CNN is largely used for the recognition of images. The CNN is a special form of multi-layer NN. Like other networks, CNNs also make the use of back propagation algorithms for training, and the distinction lies in their architectures \cite{Lecun1998} \cite{simard2003best}. The CNN is very well suited to represent the structure of an image, as there is a strong relationship of image pixels to their neighbouring pixels and have very small correlation with far away pixels. Furthermore, the CNN's strategy of weight sharing ensures that the similar properties as texture and brightness can be shared by different parts of an image. CNN can efficiently extract and abstract 2D features. The shape variations can be effectively absorbed by the CNN max-pooling layer. Further, the involvement of CNN with less parameters than a similar sized fully connected network has been made possible by sparse connection with tied weights. Most considerably, the gradient-based learning algorithm can train the CNN and the CNN suffers less from the diminishing gradient problem. The recognition of text using CNN is more tricky and tough task than image recognition since the characters and words can have different appearances according to distinct writers, writing styles and writing surfaces. Simply by employing DL techniques, the promising results can be attained for pattern recognition. Nonetheless, with the intention of getting best results in the area of pattern recognition using DL, there are some other challenges that need to be dealt for DL as it is always required to choose the most appropriate DL framework for pattern recognition. As an illustration, for handwriting recognition, CNN was initially used in digits recognition \cite{lecun1998gradient}. Then CNN and its variants are progressively used in various other handwriting recognition applications. The CNN is the most successful model for image analysis. Since late seventies, the work on CNNs has been done \cite{fukushima1982neocognitron} and these were already applied for image analysis in medical field in 1995 \cite{lo1995artificial}. The CNNs were first successfully applied in real-world application in LeNet \cite{lecun1998gradient} for recognition of handwritten digits. Regardless of CNNs initial achievement, its use did not get momentum until several new techniques were developed to train deep networks efficiently, and advancements were made in core computing systems. The watershed was a contribution \cite{krizhevsky2012imagenet} to the ImageNet challenge in 2012 and the CNN AlexNet won that competition by a huge margin. In recent years, further development has been done using related but deeper architectures \cite{russakovsky2015imagenet}. Deep convolution networks have become a technique of choice in computer vision.

\subsubsection{CNN Architecture}
\par
\begin{figure}
	\includegraphics[width=\textwidth]{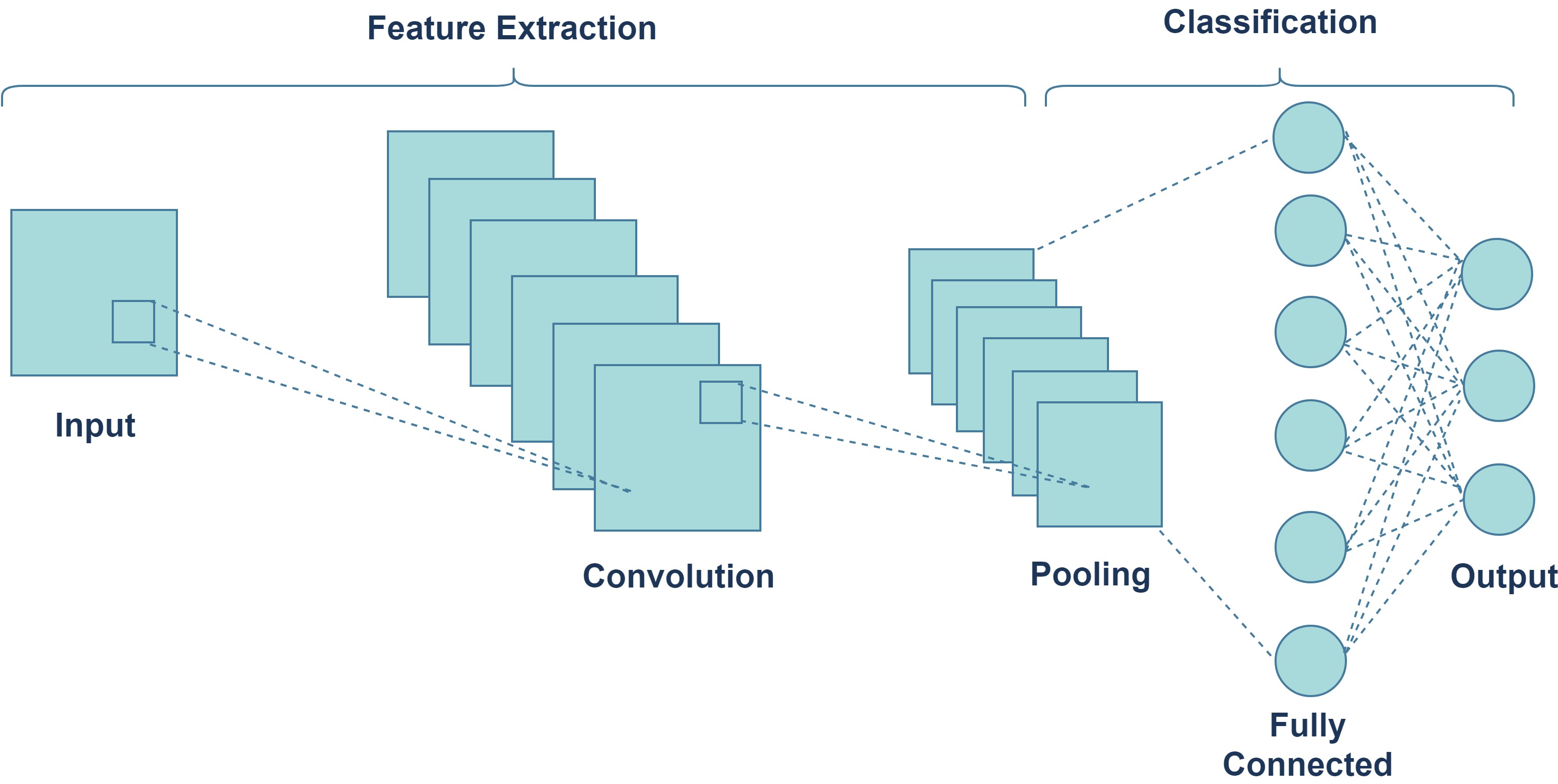}
	\caption{Basic architecture of CNN}
	\label{CNNfig1}       % Give a unique label
\end{figure}

 The architecture of CNN comprises two main components as extraction of features and classification. To extract features, CNN's each layer obtains the output from its immediately preceding layer as an input and then it delivers the current output as an input to the instant next layer, where classifier builds the expected outputs associated with the input data. The figure \ref{CNNfig1} presents the basic architecture of CNN. Generally, the architecture of CNN has two fundamental layers as: convolution layer and pooling layer \cite{lecun1998gradient}. Convolution layer's each node carries out the convolution operation on the input nodes and does feature extraction from input. The max-pooling layer performs feature extraction using average/maximum operation on input nodes. The output of ${{n-1}^{th}}$ layer is used as an input to ${{n}^{th}}$ layer, where the inputs go through kernels set trailed by ReLU the nonlinear function. The advanced architectures of CNN make use of a stack of convolutional layers and max-pooling layers followed by completely connected and softmax layer at the end. Largely an efficient architecture of the CNN can be built using basic components as convolution layer, pooling layer, softmax layer and fully connected layer.
\subsection{Recurrent neural network}
\par The RNN is a type of DL networks in which connections in nodes form a directed graph along a temporal sequence and it allows the demonstration of temporal dynamic behaviour. Feed forward NN is also used to derive an RNN and variable length sequences of inputs can be processed by using the internal state of an RNN. This feature of an RNN makes it applicable to various unsegmented tasks as speech recognition \cite{Sak2014} \cite{Li2015RNN} and connected handwriting recognition \cite{Graves2009LSTM}. An RNN term refers to two major network classes having similar general structure, in which one RNN is a directed acyclic graph and a strictly feed forward neural network can unroll and replace it, and the second RNN is also a directed cyclic graph but it cannot be unrolled. Both of these RNNs can have extra stored states, and the storage can be directly controlled by an RNN. The storage can also be replaced by other networks or graphs, if it includes time delays or has feedback loops. These controlled states are called gated states and memories, and these are also a part of LSTMs and gated recurrent units.

\subsubsection{RNN Architecture}
\begin{figure}
	\includegraphics[width=\textwidth]{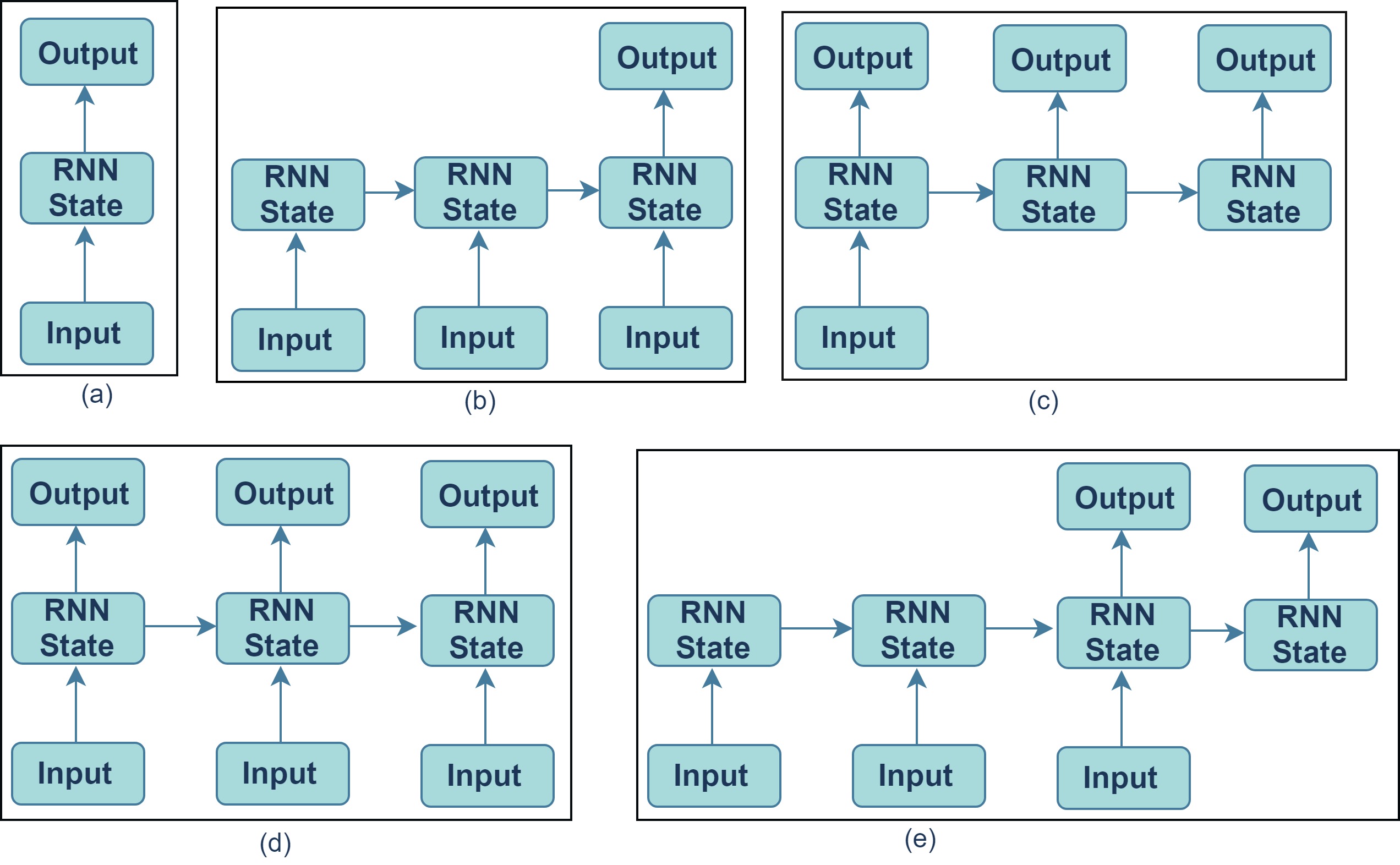}
	\caption{The different architectures of RNN (a) One to one (b) Many to one (c) One to many (d) Many to many (e) Many to many.}
	\label{RNNarchitectures}       % Give a unique label
\end{figure}

The RNNs are distinctive by the operations over a sequence of vectors over time are permitted by them. There are different architectures of RNN with respective to the applications. The figure \ref{RNNarchitectures} represents various architectures of RNN. These architectures can be categorized as: one to one, one to many, many to one and many to many.\\ \newline
\textbf{One to one:} It is a standard mode to classify without RNN and mostly used in image classification.\\
\textbf{One to many:} It takes an input and gives a sequence of outputs, and it has been successfully used in image captioning problems where a set of words output is required for a single image input.\\
\textbf{Many to one}: It has a sequence of inputs and gives one output only. It is most commonly used in those problems where inputs are in the form of sentences or words set and output is a positive/negative expression.\\
\textbf{Many to many:} It produces a sequence of outputs for a sequence of inputs and it is most commonly used in machine translation and video classification problems. \\ \par In machine translation problems, a sequence of words in one language is given as an input to a machine and translated to a sequence of words in other language. In video classification problems, video frames are taken as an input and each frame of the video is labelled as an output. \\

\subsubsection{Long Short Term Memory}
\par An LSTM \cite{HochreiterLSTM1997} \cite{Graves2009LSTM} is a particular form of RNN architecture that is specially created to overcome the problem of vanishing gradient \cite{Bengio1994}. An LSTM hidden layer comprises the recurrently connected memory blocks, in which each block has one or more recurrently connected memory cells, and three multiplicative gates (input, output, and forget gates) are used to activate and control it. These three gates make it feasible to store and access the contextual information over a long time period. More particularly, the activation of cell is not overwritten by new inputs until the input gate is closed. Similarly, as long the output gate remains open, the cell activation is accessible to the rest network and the forget gate controls the recurrent connection of the cell. Like CNN architecture, every LSTM layer can have multiple forward and backward layers; multiple feature maps at the output layer; and max-pooling sub-sampling is used to stack multiple LSTM layers.

\subsubsection{LSTM Architecture}
\par LSTM is an RNN architecture that takes into account the values over arbitrary intervals. LSTM is suitable for classification, processing and prediction of time series given time lags of unknown duration. Back propagation through time training algorithm is used to update weights in LSTM.
\par In last few years, there have been various advanced approaches developed for LSTM. The figure \ref{LSTMfig1} shows the diagram of LSTM. The main scheme for LSTM is the cell state called gates. LSTM can add and remove information to the gates. An input gate, output gate and forget gate are defined as following:\vspace{1mm}

\begin{figure}
	\includegraphics[width=.90\textwidth]{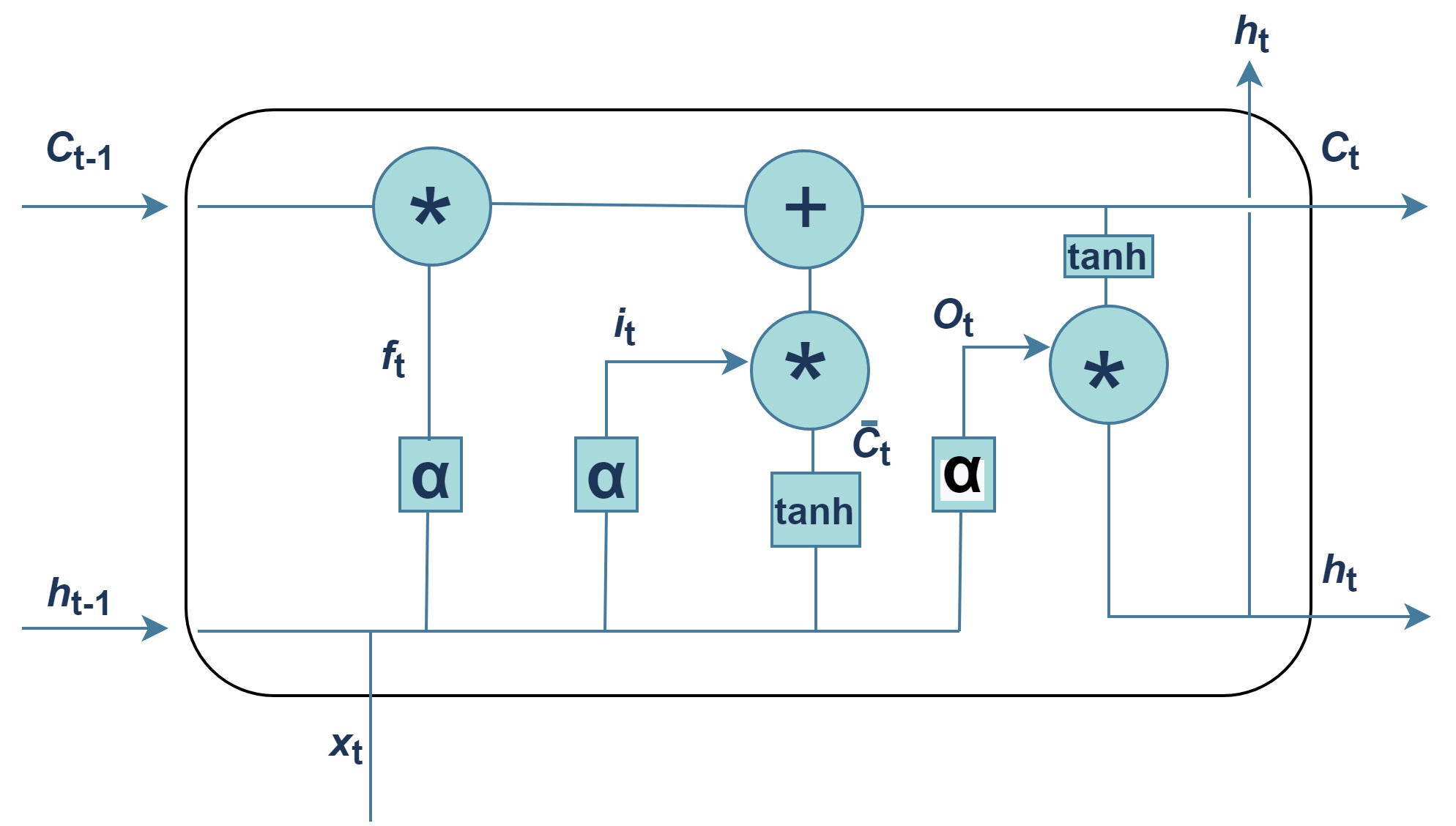}
	\caption{Diagram for LSTM}
	\label{LSTMfig1}       % Give a unique label
\end{figure}
\begin{center}
	${{f_{t}}}$ $=$ $\alpha$ $($ ${{W_{f}}}.$ $[$ ${{h_{t-1}}}$$,$ ${{x_{t}}}$ $]+$ ${{b_{f}}}),$\\
	${{i_{t}}}$ $=$ $\alpha$ $($ ${{W_{i}}}.$ $[$ ${{h_{t-1}}}$$,$ ${{x_{t}}}$ $]+$ ${{b_{i}}}),$\\
	$\tilde{{{C_{t}}}}$ $=$ $\tanh$ ${{W_{C}}}.$ $[$ ${{h_{C-1}}}$$,$ ${{x_{t}}}$ $]+$ ${{b_{C}}}),$\\
	${{{C_{t}}}}$ $=$ ${{f_{t}}}*$ ${{C_{t-1}}}$ $+$ ${{i_{t}}}*$ $\tilde{{{C_{t}}}},$\\
	${{O_{t}}}$ $=$ $\alpha$ $($ ${{W_{O}}}.$ $[$ ${{h_{t-1}}}$$,$ ${{x_{t}}}$ $]+$ ${{b_{O}}}),$\\
	${{{h_{t}}}}$ $=$ ${{O_{t}}}*$ $\tanh$ $({{C_{t}}}).$\\
\end{center}

\begin{figure}
	\includegraphics[width=0.90\textwidth]{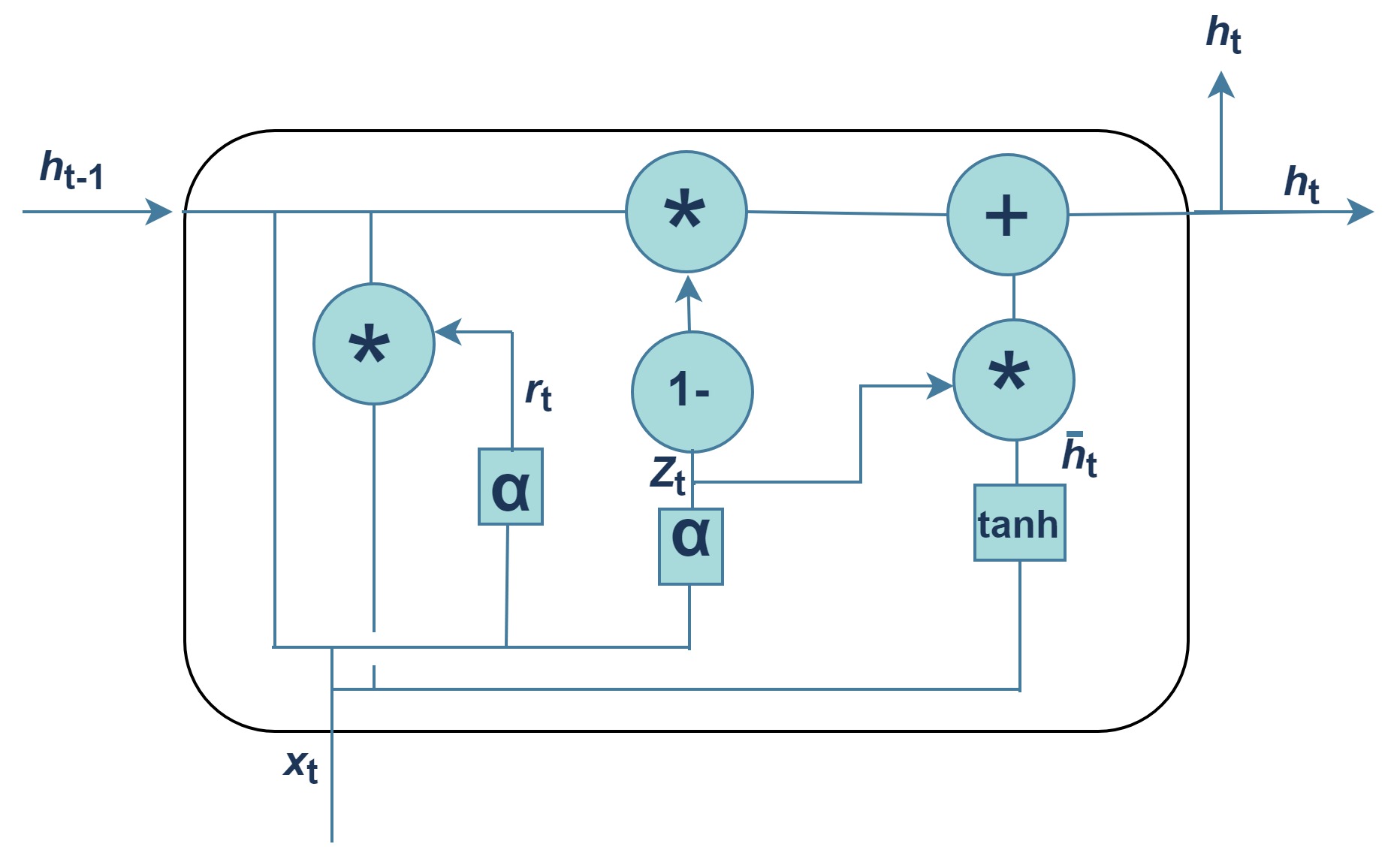}
	\caption{Gated Recurrent Unit}
	\label{GRUfig1}       % Give a unique label
\end{figure}

\vspace{1mm}
LSTM models are very well accepted for processing of temporal information. There is also little modified version of network with peephole connections \cite{gers2000recurrent}. Gated recurrent unit (GRU) comes from more variation of LSTM \cite{chung2014empirical}. GRUs are very popular among those people who work with recurrent networks. The major reason behind the recognition of GRU is its less computation cost and simple model as shown in figure \ref{GRUfig1}. GRU model is also faster as it needs fewer network parameters. But LSTM provides better results when we have enough data and computational power [174]. So, in term of computation cost, topology and complexity, GRUs are lighter versions of RNN approaches than standard LSTM. The GRU model combines the input and forget gates into a single update gate and unites the cell state and hidden state with other changes. The GRU can be expressed as following:

\begin{center}
	${{z_{t}}}$ $=$ $\alpha$ $($ ${{W_{z}}}.$ $[$ ${{h_{t-1}}}$$,$ ${{x_{t}}}$ $]),$\\
	${{r_{t}}}$ $=$ $\alpha$ $($ ${{W_{r}}}.$ $[$ ${{h_{t-1}}}$$,$ ${{x_{t}}}$ $]),$\\
	$\tilde{{{h_{t}}}}$ $=$ $\tanh$ $({{W}}.$ $[$ ${{r_{t}}}$$*$ ${{h_{t-1}}},$ ${{x_{t}}}$ $])$$,$\\
	${{{h_{t}}}}$ $=$ $(1-{{z_{t}}})$$*$${h_{t-1}}$ $+$ ${z_{t}}$$*\tilde{{{h_{t}}}}.$\\ 
\end{center}

\subsubsection{Bidirectional Long Short Term Memory}
\par In most problems of pattern recognition, it is required to access the previous and future contexts at the same time. For instance, in all problems of handwriting recognition, character recognition can be performed by recognizing the characters which appear both to the left and right of it. Further the bidirectional RNNs (BRNNs) \cite{Schuster1997BRNN} can be employed to attain the context information in left and right directions along the input sequence. In two different hidden layers of BRNNs, one layer is employed for processing of input sequence in forward direction and the subsequent in backward direction. Since both the hidden layers of BRNN are connected to the same layer of output, so the access of past and future context of every point in the sequence is given by it. BRNNs were effectively used to predict protein structure and speech processing \cite{Schuster1997BRNN}, and BRNNs did better than the standard RNNs in different tasks of sequence learning. BLSTM is a combination of LSTM and BRNN.
\subsubsection{BLSTM Architecture}
\par
\begin{figure}
	\includegraphics[width=1\textwidth]{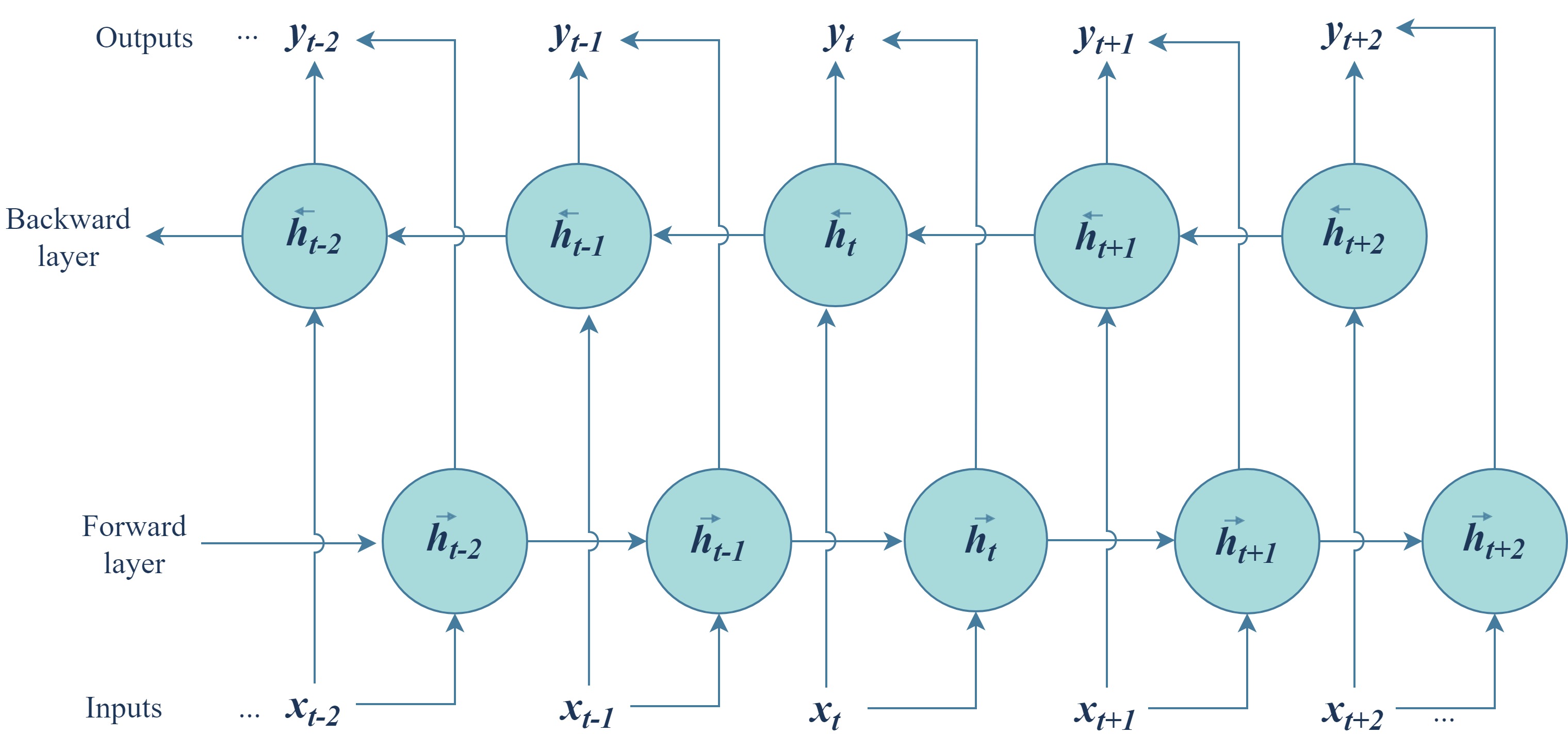}
	\caption{BLSTM architecture}
	\label{BLSTMfig1}       % Give a unique label
\end{figure}

 Bidirectional LSTMs process the input sequences in both directions having two sub layers for consideration of the full input context. The figure \ref{BLSTMfig1} presents the architecture of BLSTM. Two sub layers of BLSTM can compute both forward ($\overrightarrow{h}$) and backward ($\overleftarrow{h}$) hidden layers. Both $\overrightarrow{h}$ and $\overleftarrow{h}$ are combined for the computation of the output sequence ($y$) as following:

\begin{center}
	${\overrightarrow{h}}_t$ = $\mathcal{H}(W_{x\overrightarrow{h}}x_{t}$ $+$ $W_{\overrightarrow{h}\overrightarrow{h}}\overrightarrow{h}_{t-1}$ $+$ $b_{\overrightarrow{h}})$\\
	${\overleftarrow{h}}_t$ = $\mathcal{H}(W_{x\overleftarrow{h}}x_{t}$ $+$ $W_{\overleftarrow{h}\overleftarrow{h}}\overleftarrow{h}_{t+1}$ $+$ $b_{\overleftarrow{h}})$\\
	${y}_t$ = $(W_{\overrightarrow{h}y}\overrightarrow{h}_{t}$ $+$ $W_{\overleftarrow{h}y}\overleftarrow{h}_{t}$ $+$ $b_{\overleftarrow{h}})$\\
\end{center}

\section{Results of CNN and RNN}\label{sec4}

\subsection{Results of CNN in Handwriting Recognition}
\par With rapid development of computation techniques, the GPU-accelerated computing techniques have been exploited to train CNNs more efficiently. Nowadays, CNNs have already been successfully applied to handwriting recognition, face detection, behaviour recognition, speech recognition, recommender systems, image classification, and NLP. In deep learning, although CNN classification technique of neural systems is mostly used in image classification or image data, but it has attained good results for handwriting recognition too. The table \ref{tab1_HWR_results} presents the selected handwriting recognition results using CNN.  

\begin{longtable}{p{0.12\textwidth}p{0.15\textwidth}p{0.30\textwidth}p{0.20\textwidth}p{0.12\textwidth}}
	\caption{Results of CNN in Handwriting Recognition}
	\label{tab1_HWR_results}\\
	\hline\noalign{\smallskip}
	\textbf{Script}  &  \textbf{Reference}  &  \textbf{Methodology}  &  \textbf{Dataset}  & \textbf{Accuracy}\\  
	\hline
	Kannada digits & Gu \cite{EmilyXiaoxuanGu2021} & CNN based architecture & Kannada-MNIST (test set) & 98.77\%\\
	Roman digits & Gupta and Bag \cite{gupta2021cnn} & CNN(2 layers) & MNIST & 99.68\%\\
	Devanagari digits & Gupta and Bag \cite{gupta2021cnn} & CNN(2 layers) & CMATERdb 3.2.1 & 97.56\%\\
	Bangla digits & Gupta and Bag \cite{gupta2021cnn} & CNN(2 layers) & CMATERdb 3.1.1 & 96.35\$\\
	Telugu digits & Gupta and Bag \cite{gupta2021cnn} & CNN(2 layers) & CMATERdb 3.4.1 & 98.82\%\\
	Arabic digits & Gupta and Bag \cite{gupta2021cnn} & CNN(2 layers) & CMATERdb 3.3.1 & 96.53\%\\
	Odia digits & Gupta and Bag \cite{gupta2021cnn} & CNN(2 layers) & ISI-Odia & 97.76\%\\
	Devanagari digits & Gupta and Bag \cite{gupta2021cnn} & CNN(2 layers) & ISI-Devanagari & 98.31\%\\
	Bangla digits & Gupta and Bag \cite{gupta2021cnn} & CNN(2 layers) & ISI-Bangla & 96.70\%\\
	Gujarati digits & Gupta and Bag \cite{gupta2021cnn} & CNN(2 layers) & Gujarati dataset & 99.22\%\\
	Punjabi digits & Gupta and Bag \cite{gupta2021cnn} & CNN(2 layers) & Punjabi dataset & 99.43\%\\	
	Roman digits & Kusetogullari et al. \cite{kusetogullari2020ardis} & CNN & ARDIS & 98.60\%\\
	Kannada digits & Gati et al. \cite{Gati2019} & CNN based architecture with skip connections & Kannada-MNIST(Dig-MNIST) & 85.02\%\\
	Gurmukhi strokes & Singh et al. \cite{singh2020self} & CNN, self controlled RDP based features & Gurmukhi & 94.13\%\\
	Roman digits & Singh et al. \cite{singh2020self} & CNN, self controlled RDP based features & UNIPEN & 93.61\%\\	
	Kannada digits & Prabhu \cite{prabhu2019kannadamnist} & End-to-end training using CNN based architecture & Kannada-MNIST (test set) & 96.80\%\\				
	Malayalam characters & Manjusha et al. \cite{MANJUSHA2019637} & CNN based on scattering transform-based wavelet filters as feature extractor and Linear SVM as classifier & Amrita\_MalCharDb & 91.05\\
	Roman digits & Chowdhury et al.\cite{chowdhury2019bangla} & CNN based architecture & MNIST & 99.25\%\\
	Bangla characters & Chowdhury et al.\cite{chowdhury2019bangla} & CNN based architecture & Banglalekha-isolated & 91.81\%\\
	Bangla characters & Chowdhury et al.\cite{chowdhury2019bangla} & CNN based architecture & Ekush & 95.07\%\\
	Bangla characters & Chowdhury et al.\cite{chowdhury2019bangla} & CNN based architecture & CMATERdb 3.1.2 & 93.37\%\\		
	Bangla numerals & Gupta et al. \cite{gupta2019multiobjective}  & Multi-objective optimisation to find the informative regions of character image + CNN features & Isolated handwritten Bangla numerals & 96.54\%\\
	
	Bangla characters & Gupta et al. \cite{gupta2019multiobjective}  & Multi-objective optimisation to find the informative regions of character image + CNN features & Isolated handwritten Bangla basic characters & 85.19\%\\				
	English numerals & Gupta et al. \cite{gupta2019multiobjective}  & Multi-objective optimisation to find the informative regions of character image + CNN features & Isolated handwritten English numerals & 97.87\%\\		
	Devanagari characters & Gupta et al. \cite{gupta2019multiobjective}  & Multi-objective optimisation to find the informative regions of character image + CNN features & Isolated handwritten Devanagari characterss & 87.23\%\\		
	Roman digits & Chakraborty et al. \cite{chakraborty2019feature} & feature map reduction in CNN & MNIST & 99.19\%\\
	Roman digits & Arora and Bhatia \cite{Arora2018HWR_DL} & CNN, Keras & MNIST & 95.63\%, 99.20\%\\
	Malayalam characters & Manjusha et al. \cite{manjusha2018integrating} & CNN based on scattering transform-based wavelet filters (ScatCNN) & Malayalam\_DB & 93.77\%\\
	Roman digits & Manjusha et al. \cite{manjusha2018integrating} & ScatCNN & MNIST & 99.31\%\\
	Bangla numerals & Manjusha et al. \cite{manjusha2018integrating} & ScatCNN & ISI & 99.22\%\\
	Chinese characters & Manjusha et al. \cite{manjusha2018integrating} & ScatCNN & CASIA HWDB1.1 & 92.09\%\\
	English words & Kang et al. \cite{kang2018convolve} & Attention based sequence to sequence model & IAM & 82.55\%\\
	English characters & Kang et al. \cite{kang2018convolve} & Attention based sequence to sequence model & IAM & 93.12\%\\
	Roman digits & Sarkhel et al. \cite{SARKHEL201778} & A multi-column multi-scale CNN architecture + SVM & CMATERdb 3.4.1 & 99.50\%\\
	Roman digits & Sarkhel et al. \cite{SARKHEL201778} & A multi-column multi-scale CNN architecture + SVM & MNIST & 99.74\%\\

	English(mostly) words & Poznanski and Wolf \cite{Poznanski_2016_CVPR} & CNN-N-Gram & IAM & 93.55\%\\
	English(mostly) characters & Poznanski and Wolf \cite{Poznanski_2016_CVPR} & CNN-N-Gram & IAM & 96.66\%\\
	
	French words & Poznanski and Wolf \cite{Poznanski_2016_CVPR} & CNN-N-Gram & RIMES & 93.10\%\\
	French characters & Poznanski and Wolf \cite{Poznanski_2016_CVPR} & CNN-N-Gram & RIMES & 98.10\%\\
	
	Hangual & Kim and Xie \cite{Kim2015HandwrittenHangual} & Deep convolutional neural netwok (DCNN) & SERI95a and PE92 & 95.96\%, 92.92\%\\
	Roman digits & Wan et al. \cite{wan2013regularization} & CNN based architecture with DropConnect layer & MNIST & 99.79\%\\
	Roman digits & Krizhevsky et al. \cite{krizhevsky2012imagenet} & CNN, LeNet-5 system & MNIST & 99.10\%\\						
	\hline
\end{longtable}

One of the classical models of CNN was the LeNet-5 system. Its accuracy rate on MNIST data-set was above 99\%. It was extensively used for identification of handwritten checks on banks, but it could not recognize large images. With the advancement of technology, Graphics Processing Unit (GPU) was developed, then in 2012, Krizhevsky et al. \cite{krizhevsky2012imagenet} employed an efficient GPU supported program for solving ImageNet problem, which also made CNN application popular. Actually, one of the problems of using CNN was that it took much time to train the network because of the many hidden nodes in the network. But the GPUs' faster parallel computing, overcame this problem too. A CNN based architecture with DropConnect layer was proposed by Wan et al. \cite{wan2013regularization} in 2013, where they attained 99.79\% recognition accuracy for MNIST digits. DropConnect generalizes Hinton et al.'s Dropout \cite{hinton2012improving} to the complete connectivity structure of a fully connected neural network (NN) layer. They provided both empirical results and theoretical justification for showing that DropConnect helps to regularize large NN models. As deep convolutional neural netwok (DCNN) comprises many layers, so it can model much more complicated functions than shallow networks. Motivating from DCNN's great results in various machine learning and pattern recognition problems, in 2015, Kim and Xie \cite{Kim2015HandwrittenHangual} developed a new recognizer based on deep CNN to improve the Hangual handwrititng recognition performance. They built their own Hangul recognizers based on DCNN and developed various novel techniques for performance and networks training speed improvement. They evaluated their proposed recognizers on image datasets of Hangul, named SERI95a and PE92, where recognition results as 95.96\% on SERI95a and 92.92\% on PE92 are attained. In 2016, Poznanski and Wolf \cite{Poznanski_2016_CVPR} proposed a CNN-N-Gram based system for handwriting recognition, and recognized handwritten English and French words with 93.55\% and 93.10\% accuracy, respectively. Different persons' variation in writing styles and single person's variation in handwriting from time to time, make recognition of the local invariant patterns of a handwritten digit and character difficult. For this purpose, in 2017, Sarkhel et al. \cite{SARKHEL201778} proposed a non-explicit feature based approach, specifically it was a multi-column multi-scale CNN based architecture. Their proposed approach has been validated on different datasets of isolated handwritten digits and characters of Indic scripts, and best results are attained on MNIST dataset that is 99.74\% without any data augmentation to the original dataset. Inspired from the deep learning's role in image classification, in 2018, Arora and Bhatia \cite{Arora2018HWR_DL} used Keras for classification of handwritten images of MNIST dataset. In fact, they used feed forward NN and CNN to extract features and training the model, it used Stochastic Gradient Descent for optimization. In their work, for classification of handwritten digits, it is observed that CNN attained greater accuracy in comparison to feed forward, and CNN obtained 95.63\% and 99.20\% accuracy for 5 and 20 iterations, respectively. Malayalam handwritten character recognition is very challenging, due to the isomorphic nature of character classes and a large number of character classes. To recognize handwritten Malayalam characters, in 2018, Manjusha et al. \cite{manjusha2018integrating} replaced the convolutional feature maps of first layer in CNN architecture with scattering transform-based feature maps, and attained 93.77\% as recognition accuracy. Scattering transform can compute stable invariant description of input patterns where it applies a series of wavelet decomposition, modulus and averaging operations. Their proposed hybrid CNN \cite{manjusha2018integrating} also achieved above 99\% recognition accuracy for MNIST digits and ISI Bangla numerals datasets. A convolve, attend and spell, an attention based sequence to sequence model to recognize handwritten words without the use of HTR system's traditional components, as connectionist temporal classification, language model nor lexicon was presented by Kang et al. \cite{kang2018convolve} in 2018. It was an end-to-end system that contained an encoder, decoder and attention mechanism, and it outperformed most of the existing best results, and it attained 93.12\% character recognition accuracy and 82.55\% word recognition accuracy for IAM dataset on word-level. In 2019, Chowdhury et al. \cite{chowdhury2019bangla} used CNN to develop a handwritten character recognition model, and attained 99.25\% accuracy for MNIST digits and 91.81\% accuracy for Banglalekha-isolated characters. In 2019, Gupta et al. \cite{gupta2019multiobjective} proposed an opposition based multi-objective optimisation search algorithm to find the informative regions of character images, where they also used CNN features to evaluate the proposed work on different Indic scripts' isolated units of handwriting and obtained good results for isolated Bangla basic characters, Bangla numerals, English numerals, and isolated Devanagari characters. Considering the research efforts for Malayalam character handwriting recognition, Manjusha et al. \cite{MANJUSHA2019637} developed a handwritten character image database of Malayalam language script in 2019. In their work, recognition experiments were conducted by using different techniques of feature extraction. Among the used feature descriptors, scattering CNN attained the best recognition accuracy of 91.05\%. In 2019, Prabhu \cite{prabhu2019kannadamnist} created a new dataset for handwritten digits of Kannada language, which is called Kannada-MNIST dataset, and attained best results as 96.80\% using CNN based architecture. In 2019, Gati et al. \cite{Gati2019} described how great results and performance can be attained on a very challenging Dig-MNIST dataset using a custom-built model based on the skip CNN architecture, where 85.02\% recognition accuracy was attained using proposed model trained on Kannada-MNIST and tested on the Dig-MNIST dataset without any pre-processing. In 2019, Chakraborty et al. \cite{chakraborty2019feature} proposed for reduction of the feature maps which are used in training the CNN for reduction of computation time and storage space. Experimental results proved that the time requirement for training the CNN decreased with reduction in number of feature maps without affecting the accuracy much, and above 99\% accuracy rate was attained for MNIST digit dataset. In 2020, Kusetogullari et al. \cite{kusetogullari2020ardis} introduced different datasets of digits in ARDIS, and attained best recognition accuracy for digits using CNN that is 98.60\%. A novel self-controlled RDP point based smaller size feature vector approach to recognize online handwriting was proposed by Singh et al. \cite{singh2020self} in 2020, where they employed a CNN based network that trains in a few minutes on a single machine without GPUs due to the use of Conv1Ds, and it attained 94.13\% and 93.61\% recognition rates for Gurmukhi and UNIPEN datasets, respectively. Recently, in 2021, Gu \cite{EmilyXiaoxuanGu2021} proposed a CNN based model to classify the Kannada-MNIST dataset and made analysis of the proposed model performance on training, testing and validation sets. The CNN model was trained on more than 51000 images and it was validated over 9000 images for 30 epochs, where the CNN model attained a testing accuracy of 98.77\%, and it outperformed other methods as SVM, logistic regression and a CNN baseline. This study is the evidence for the capability of proposed CNN model, and it also demonstrates the benefit of using a CNN architecture over other classification methods when performing handwritten character recognition jobs. A script independent CNN based system to recognize numerals was developed by Gupta and Bag \cite{gupta2021cnn} in 2021, it is a system to recognize handwritten digits written in multi languages and it is independent of fusion where it has just 10 classes corresponding to every numeric digit. This was the first study that addressed the problem of multilingual numerals recognition, where experimental results attained the accuracy of 96.23\% for eight Indic scripts collectively. The attained results are promising and demonstrates the hypothesis that multilingual handwritten numeral recognition is void with CNN.
\subsection{Results of RNN in Handwriting Recognition}
RNNs are very powerful machine learning models and have found use in a wide range of areas where sequential data is dealt. RNNs have been widely used in prediction problems, machine translation, face detection, speech Recognition, OCR based image recognition and handwriting recognition etc. RNNs have received great success when working with sequential data, generally in the field of handwriting recognition. The table \ref{tab2_HWR_results} presents the selected handwriting recognition results using RNN.

\begin{longtable}{p{0.12\textwidth}p{0.15\textwidth}p{0.30\textwidth}p{0.20\textwidth}p{0.12\textwidth}}
	\caption{Results of RNN in Handwriting Recognition}
	\label{tab2_HWR_results}\\
	\hline\noalign{\smallskip}
	\textbf{Script} & \textbf{Reference} & \textbf{Methodology} & \textbf{Dataset} &\textbf{Accuracy}\\  
	\hline	
	English words&Pham et al. \cite{pham2014dropout}&LSTM with dropout at the topmost hidden layer&IAM&60.52\%\\
	English words&Pham et al. \cite{pham2014dropout}&LSTM with dropout at multiple layers&IAM&68.56\%\\
	French words&Pham et al. \cite{pham2014dropout}&LSTM with dropout at the topmost hidden layer&Rimes&63.97\%\\
	French words&Pham et al. \cite{pham2014dropout}&LSTM with dropout at multiple layers&Rimes&72.99\%\\			
	English characters&Pham et al. \cite{pham2014dropout}&LSTM with dropout at the topmost hidden layer&IAM&81.55\%\\
	English characters&Pham et al. \cite{pham2014dropout}&LSTM with dropout at multiple layers&IAM&86.08\%\\
	French characters&Pham et al. \cite{pham2014dropout}&LSTM with dropout at the topmost hidden layer&Rimes&87.83\%\\
	French characters&Pham et al. \cite{pham2014dropout}&LSTM with dropout at multiple layers&Rimes&91.38\%\\		
	English words&Doetsch et al. \cite{doetsch2014fast}&LSTM-RNN&IAM&87.80\%\\
	French words&Doetsch et al. \cite{doetsch2014fast}&LSTM-RNN&IAM&87.10\%\\
	Bangla characters&Chollet et al. \cite{chollet2015keras}&LSTM&Banglalekha-isolated&87.41\%\\	
	Bangla characters&Chollet et al. \cite{chollet2015keras}&LSTM&Ekush&93.06\%\\			
	Arabic words&Chherawala et al. \cite{Chherawala2016}&Weighted Vote Combination of RNN&FN/ENIT&96\%\\
	French words&Chherawala et al. \cite{Chherawala2016}&Weighted Vote Combination of RNN&RIMES&95.2\%\\
	English words&Shkarupa et al. \cite{shkarupa2016offline}&CTC+BLSTM&handwritten medieval Latin text&78.10\%\\
	English words&Shkarupa et al. \cite{shkarupa2016offline}&Sequence to sequence+LSTM&handwritten medieval Latin text&72.79\%\\     
	English Words&Wigington et al. \cite{Wigington2017}&RNN+CTC&IAM&80.93\%\\
	French Words&Wigington et al. \cite{Wigington2017}&RNN+CTC&Rimes&88.71\%\\
	English characters&Wigington et al. \cite{Wigington2017}&RNN+CTC&IAM&93.93\%\\
	French characters&Wigington et al. \cite{Wigington2017}&RNN+CTC&Rimes&96.91\%\\
	English words&Dutta et al. \cite{dutta2018improving}&Hybrid CNN-RNN network&IAM&87.39\%\\
	English characters&Dutta et al. \cite{dutta2018improving}&Hybrid CNN-RNN network&IAM&95.12\%\\
	French words&Dutta et al. \cite{dutta2018improving}&Hybrid CNN-RNN network&RIMES&92.96\%\\
	French characters&Dutta et al. \cite{dutta2018improving}&Hybrid CNN-RNN network&RIMES&97.68\%\\
	English words&Dutta et al. \cite{dutta2018improving}&Hybrid CNN-RNN network&GW&87.02\%\\
	English characters&Dutta et al. \cite{dutta2018improving}&Hybrid CNN-RNN network&GW&95.71\%\\		
	English words&Krishnan et al. \cite{krishnan2018word}&Convolutional recurrent neural network (CRNN)&IAM&94.90\%\\
	English words&Sueiras et al. \cite{SUEIRAS2018119}&Sequence to
	sequence NN&IAM&87.30\%\\
	French words&Sueiras et al. \cite{SUEIRAS2018119}&Sequence to
	sequence NN&IAM&93.40\%\\		
	English characters&Krishnan et al. \cite{krishnan2018word}&CRNN&IAM& 97.44\%\\
	Bengali words&Ghosh et al. \cite{GHOSH2019203RNNHW}&BLSTM&Bengali dataset of 120000 words&95.24\% (lexicon 1K)\\			
	Bengali words&Ghosh et al. \cite{GHOSH2019203RNNHW}&BLSTM&Bengali dataset of 120000 words&90.78\% (lexicon 5K)\\
	Bengali words&Ghosh et al. \cite{GHOSH2019203RNNHW}&BLSTM&Bengali dataset of 120000 words&87.38\% (lexicon 10K)\\    			
	Devanagari words&Ghosh et al. \cite{GHOSH2019203RNNHW}&BLSTM&Bengali dataset of 120000 words&99.50\% (lexicon 1K)\\			
	Devanagari words&Ghosh et al. \cite{GHOSH2019203RNNHW}&BLSTM&Bengali dataset of 120000 words&96.27\% (lexicon 5K)\\			
	Devanagari words&Ghosh et al. \cite{GHOSH2019203RNNHW}&BLSTM&Bengali dataset of 120000 words&94.34\% (lexicon 10K)\\			
	English words&Geetha et al. \cite{Geetha2021effective}&CNN-RNN&IAM&95.20\%\\
	English characters&Geetha et al. \cite{Geetha2021effective}&CNN-RNN&IAM&97.48\%\\
	French words&Geetha et al. \cite{Geetha2021effective}&CNN-RNN&RIMES&98.14\%\\
	French characters&Geetha et al. \cite{Geetha2021effective}&CNN-RNN&RIMES&99.35\%\\		
	\hline
\end{longtable}

\par In 2014, Pham et al. \cite{pham2014dropout} presented that the dropout can improve the performance of RNN greatly. The word recognition networks having dropout at the topmost layer improved the character and word recognition by 10\% to 20\%, and when dropout used with multiple LSTM layers, then it further improved the performance by 30\% to 40\%. They reported the best results on Rimes dataset as 91.38\% and 72.99\% for character and word recognition, respectively. The simple RNN was modified by Koutnik et al. \cite{koutnik2014clockwork} in 2014, and introduced a powerful Clockwork RNN (CW-RNN), where hidden layers were divided into different modules and every layer processed the inputs at its own temporal granularity, which made computations over prescribed clock rate only. The CW-RNN reduced the number of parameters of simple RNN, and also improved the speed and performance significantly. For online handwriting recognition, CW-RNN outperformed the simple RNN and LSTM, and improved recognition accuracy by 20\% for English sentences. A modified topology for LSTM-RNN that controlled the shape of squashing functions in gating units was demonstrated by Doetsch et al. \cite{doetsch2014fast} in 2014. An efficient framework of mini batch training at sequence level in combination with sequence chunking approach was also proposed by them. They evaluated their framework on IAM and RIMES datasets by using GPU based implementation, and it was three times faster in training RNN models which outperformed the state-of-the-art recognition results, where 87.80\% and 87.10\% recognition accuracies were attained for handwritten words of IAM and RIMES datasets, respectively. In 2015, an image classification system using LSTM with Keras was built and it was applied to handwritten Bangla character datasets, where it achieved 87.41\% and 93.06\% recognition accuracy for two different datasets of Bangla characters \cite{chollet2015keras}. This architecture consisted of one LSTM layer having 128 units, had activation function as 'ReLU' and recurrent activation function had been set to 'hard sigmoid'. In 2016, Chherawala et al. \cite{Chherawala2016} proposed a novel method to extract the promising features of handwritten word images. They proposed a framework to evaluate feature set based on collaborative setting. In their work, they employed weighted vote combination of RNN classifiers, where particular feature set was used to train every RNN. The major contribution of their study was the quantification of the feature sets' importance through weight combination, and it also showed their complementarity and strength. They used RNN because of the state-of-the-art results, and provided the first feature set benchmark for RNN classifier. They evaluated different feature sets on different datasets of Arabic and Latin scripts, and attained best accuracies as 96\% and 95.2\% for IFN/ENIT and RIMES datasets, respectively. For historic handwritten Latin text recognition, two important approaches based on RNN were proposed by Shkarupa et al. \cite{shkarupa2016offline} in 2016. Their first approach used connectionist temporal classification (CTC) output layer, and attained 78.10\% word level accuracy. The other approach used sequence-to-sequence learning, and attained 72.79\% word level accuracy. In their work, when CTC approach was used with BLSTM, it outperformed the sequence-to-sequence based approach used with LSTM. Their proposed system of handwriting recognition considered unsegmented word images as input and provided decoded strings as output. In 2017, Wigington et al. \cite{Wigington2017} presented two data normalization and augmentation techniques, and these were used with CNN and LSTM. These techniques reduced the character error rate and word error rate significantly, and significant results were reported for handwriting recognition tasks. The novel normalization technique was applied to both word and line images. Their proposed approaches attained high accuracies for both characters and words over several existing studies, where IAM dataset character and word level recognition accuracy was reported as 96.97\% and 94.39\%, respectively. In 2018, Dutta et al. \cite{dutta2018improving} proposed a modified CNN-RNN based hybrid architecture and mainly focussed for effective training with: (a) network's efficient initialization with the use of synthetic data in pretraining, (b) slant correction with image normalization and (iii) domain specific transformation of data and distortion to learn important invariances. In their work, a detailed ablation study for analysis of the contribution of individual module was performed and the results for unconstrained line and word recognition on IAM, RIMES and GW datasets were presented at par literature, where they attained lexicon free word recognition accuracies as 87.39\%, 92.96\% and 87.02\% on these three datasets, respectively. To represent handwritten word images efficiently, an HWNet v2 architecture was presented by Krishnan et al. \cite{krishnan2018word} in 2018. The state-of-the-art attribute embedding was enabled by this work. An end-to-end embedding framework was demonstrated by it, and it used textual representation and synthetic image for learning complementary information to embed text and images. It also improved the word recognition performance using a convolutional recurrent neural network (CRNN) architecture, by using the synthetic data and spatial transformer layer, and attained character and word level accuracies on IAM dataset as 97.44\% and 94.90\%, respectively. In 2018, a system based on sequence to sequence architecture with convolutional network was proposed by Sueiras et al. \cite{SUEIRAS2018119} to recognize offline handwriting. This model had three major components, where first convolutional network extracted relevant features of the characters present in the word. Then RNN captured the sequential relationships of extracted features. Thirdly, the input word was predicted by decoding the sequence of characters with another RNN. Their proposed system was tested on handwritten words of IAM and RIMES datasets, and attained the recognition accuracy as 87.3\% and 93.6\%, respectively, where no language model was used and results were attained with closed dictionary. In 2019, Ghosh et al. \cite{GHOSH2019203RNNHW} presented a new online handwritten word recognition system based on LSTM and BLSTM versions of RNN, and recognized Devanagari and Bengali words in lexicon dependent environment with above 90\% (for lexicon size 5K) recognition accuracy. Their proposed approach divided every handwritten word into upper, middle, and lower zones horizontally, and reduces the basic stroke order variations with in a word. Further, they also used various structural and directional features of different zones' basic strokes of handwritten words. In 2021, Geetha et al. \cite{Geetha2021effective} proposed a hybrid model to recognize handwritten text by utilizing deep learning that used sequence-to-sequence approach. It used various features of CNN and RNN-LSTM. It used CNN to extract features of handwritten text images. The extracted features were then modelled with a sequence-to-sequence approach and fed in RNN-LSTM to encode the visual features and decoded the sequence of letters present in handwritten image. Their proposed model was tested with IAM and RIMES datasets, where above 95\% accuracy was attained using CNN-RNN for handwritten words of English and French.

\section{General Observations}\label{sec5}

\par In this section, an analysis of various techniques used for deep learning architectures, and deep learning use in handwriting recognition and other related fields are presented.
\begin{itemize}
	\item Deep learning has been effectively used in various emerging fields to solve complex problems of real world with different deep learning architectures. Deep learning architectures employ different activation functions for the achievement of state-of-the-art performances, to perform various computations between the hidden and output layers of deep learning architectures. Further, the advancement in deep learning architectures' configuration brings new challenges, particularly for the selection of right activation functions to perform in various domains from the classification of objects \cite{krizhevsky2012imagenet} \cite{szegedy2015going} \cite{he2015delving} \cite{md2017hyperspectral}, speech recognition \cite{sainath2015deep} \cite{graves2013speech}, segmentation \cite{badrinarayanan2017segnet} \cite{hu2018learning}, machine translation \cite{vinyals2014grammar} \cite{liu2018deep}, scene description \cite{karpathy2015deep} \cite{pinheiro2014recurrent}, weather forecasting \cite{grover2015deep} \cite{hossain2015forecasting}, cancer detection \cite{albarqouni2016aggnet} \cite{wang2016deep} \cite{cruz2013deep}, self-driving cars \cite{uccar2017object} \cite{chen2015deepdriving} and other adaptive systems. With such challenges, the comparison of present trends in the application of activation functions employed in deep learning, portrays a gap of literature in this direction.
	
	\item Deep learning performs most of the things that are familiar to machine learning approaches. Deep learning techniques can be used both in supervised learning based applications that require prediction of one or more outcomes or labels related to each data point in place of regression approaches, as well as in unsupervised learning based applications that need summary, explanation and identification of interesting patterns in a data set in the form of clustering.
	
	\item Deep learning techniques surpass the existing state of the art in various studies of healthcare like patient and disease categorization, basic biological study, genomics and treatment development.
	
	\item Despite deep learning has dominated over competing machine learning approaches in many fields and made quantitative improvements in predictive performance, deep learning has yet to solve many problems. Deep learning has not completely transformed the study of human disease. It has yet to realize its trans-formative strength and to encourage a strategic inflection point.
	
	\item Using deep learning in speech recognition, there have been great performance improvements with error rates dropped from more than 20\% to less than 6\% and exceeded human performance in the past years \cite{xiong2017achieving} \cite{saon2017english}.
	
	\item In medical imaging, diabetic macular oedema \cite{gulshan2016development}, diabetic retinopathy \cite{gulshan2016development}, skin lesion \cite{esteva2017dermatologist} and tuberculosis \cite{lakhani2017deep}, deep learning based classifiers are greatly successful and can be compared to clinical performance. Some of these areas, we have surpassed the lofty bar than others, generally, those that are more similar to the non-biomedical tasks that are now monopolized by deep learning. Deep learning can point experts to the most challenging cases that require manual review, even if the risk of false negatives must be addressed. 
	
	\item Deep learning techniques also prioritize experiments and assist discovery. As an illustration, in chemical screening for discovery of drugs, a deep learning system can successfully identify hundreds of target-specific, active small molecules from an immense search space and it would have great practical value even though its total precision is modest.
	
	\item The deep neural networks can be built resistant to the adversarial attacks. Further, there is also possibility to design reliable adversarial training methods. Thus, the findings from existing deep learning studies provide motivation for having adversarial robust deep learning models within current reach.
	
	\item Deep learning has witnessed for a great achievement of human level performance across a number of domains in biomedical science. But deep neural networks as other machine learning algorithms are also prone to errors that are also made by humans most likely, such as miss-classification of adversarial examples \cite{szegedy2013intriguing} \cite{goodfellow2014explaining}, and it can be considered that the semantics of the objects presented cannot be completely understood by these algorithms. But the alliance between deep learning algorithms and human experts addresses most of these challenges and can result in better performance than either individually \cite{wang2016deep}.
	
	\item We are confident about deep learning's future in machine learning. It is certain that the deep learning will surely revolutionize these fields, but given how rapidly these areas are evolving, we are optimistic that its full potential has not been explored yet. There are various challenges beyond improving the training and predictive accuracies. Ongoing research has begun to address most problems in these directions and proved that they are not insuperable.
	
	\item Deep learning provides a flexible way to model data in its natural form, as an illustration, molecular graphs instead of pre-computed bit vectors for drug discovery and longer DNA sequences instead of k-mers for TF binding forecasting. This kind of flexible input feature interpretations have incited creative modelling approaches that are not feasible with rest of machine learning techniques. In forthcoming years, large collections of input data can be summarized into interpretable models by deep learning algorithms, and it will encourage scientists to ask those questions which they do not know how to ask.
	
	\item Although the deep learning has yet to attain more, existing studies show that it possesses the capability of faster and more reliable results. That power may well trigger a shift away from the currently employed decision support methods, such as support vector machines and k-nearest neighbour, towards deep learning.
	
	\item Deep leaning training algorithms have a high computational complexity and it results to high run time complexity that translates into a long training time use. After choosing the architecture, there is always a need to adjust the tuning parameters. The model is influenced by both the structure selection and parameter adjustment. So, there is need to have many test runs. Reducing the deep leaning models' training phase is an active area of research. One of the challenges in deep learning is increasing the training process speed in a parallel distributed processing system \cite{chen2014big}. As the network for individual processors becomes the bottle neck \cite{najafabadi2015deep} then GPUs are used to reduce the network latency \cite{bergstra2011theano}.
	
	\item Deep learning faces certain challenges as: using deep learning for big data analysis, dealing causality in learning, scalability of approaches in deep learning, data generating ability when data does not exist for learning the system, need of energy efficient techniques for special purpose devices, learning from different domains or models together.
	
	\item Although, present deep learning models works splendidly in various applications, but the solid theory of deep learning still lacks. It is not mostly known that why and how it works essentially. It is required to make more endeavours to investigate the basic principles of deep learning. In the meantime, it is very worth on exploring how to leverage natural visual perception mechanism for further improvement in the design of deep learning models.
	
\end{itemize}

\section{Conclusion}\label{sec6}
\par The recent decade observed an increasingly rapid progress in technology, mainly backed up by the advancements in the area of deep learning and artificial intelligence. The present paper presented various architectures of deep learning and surveyed the current state-of-the-art on deep learning technologies used in handwriting recognition domain. After reviewing so many papers, the present study is able to distil the perfect deep learning methods and architectures for different handwriting recognition tasks and general observations on other related application areas too. The CNN and its derivatives are the out performers in most image analysis areas, and RNN and its derivatives are out performers in dealing with sequence data. Further, an outstanding conclusion can be drawn that the exact architecture of deep learning is an important determinant for finding a good solution in many problems. The present survey not only given a snapshot of the existing deep learning research status in handwriting recognition but also made an effort for identification of the future roadway for intended researchers. The findings indicate that there are remarkable opportunities in the deep learning research and it shows that they will not disappear anytime soon. So, the present study encourages future researchers that are interested in the area to start exploring as it currently seems to be wide open for new studies.\\

\bibliographystyle{unsrtnat}
\bibliography{mybibfile1}

\end{document}

\begin{comment}
\bibliographystyle{unsrtnat}
\bibliography{references}  %%% Uncomment this line and comment out the ``thebibliography'' section below to use the external .bib file (using bibtex) .




% 	\bibitem{keshet2016prediction}
% 	Keshet, Renato, Alina Maor, and George Kour.
% 	\newblock Prediction-Based, Prioritized Market-Share Insight Extraction.
% 	\newblock In {\em Advanced Data Mining and Applications (ADMA), 2016 12th International 
%                       Conference of}, pages 81--94,2016.

% \end{thebibliography}

\end{document}